# Solar Event Tracking with Deep Regression Networks: A Proof of Concept Evaluation

Toqi Tahamid Sarker
*Computer Science*
*Georgia State University*
Georgia, USA
tsarker3@student.gsu.edu

Juan M. Banda
*Computer Science*
*Georgia State University*
Georgia, USA
jbanda@gsu.edu

*Abstract*—With the advent of deep learning for computer vision tasks, the need for accurately labeled data in large volumes is vital for any application. The increasingly available large amounts of solar image data generated by the Solar Dynamic Observatory (SDO) mission make this domain particularly interesting for the development and testing of deep learning systems. The currently available labeled solar data is generated by the SDO mission's Feature Finding Team's (FFT) specialized detection modules. The major drawback of these modules is that detection and labeling is performed with a cadence of every 4 to 12 hours, depending on the module. Since SDO image data products are created every 10 seconds, there is a considerable gap between labeled observations and the continuous data stream. In order to address this shortcoming, we trained a deep regression network to track the movement of two solar phenomena: Active Region and Coronal Hole events. To the best of our knowledge, this is the first attempt of solar event tracking using a deep learning approach. Since it is impossible to fully evaluate the performance of the suggested event tracks with the original data (only partial ground truth is available), we demonstrate with several metrics the effectiveness of our approach. With the purpose of generating continuously labeled solar image data, we present this feasibility analysis showing the great promise of deep regression networks for this task.

## I. Introduction

In computer vision, object tracking is estimating the trajectory of a target from a set of continuous images, which is challenging since only the initial position is known. With a wide range of potential applications, such as autonomous surveillance systems, human computer interaction, and robotics, it is one of the most popular problems in computer vision. In recent years, most of the state-of-the-art object trackers [1-2] successfully applied Deep Convolutional Neural Networks (CNN) in the field of object tracking, as well as other approaches, such as multiple instance learning [3] and semi-supervised learning [4].

In the field of solar physics, the Solar Dynamic Observatory (SDO) mission by NASA captures over 70,000 high resolution images of the sun per day. The SDO Feature Finding Team (FFT) members have developed and validated detector modules that can find a wide variety of solar events [5], these detection tasks are performed with a cadence of 4 to 12 hours [6]. Therefore, the SDO mission has generated a treasure trove of data, but a large amount of it remains unlabeled, making it difficult for researchers to fully understand the trajectory of the solar events and what happens in those 'blackout' periods.

Previous attempts have been made to label the images between event detections. Applying different interpolation methods [6-7], the authors estimated the trajectories of the solar event between FFT module event reports on two known timestamps, using the events' own trajectory. This is the only known approach that aimed to fill in the gaps, but we theorize that some events might behave differently, so a more robust approach is needed.

In this work, our approach is fundamentally different since we only use the initial label as a starting point of our tracking method. Our method applies a deep regression network [2] to estimate the solar event trajectories from a set of sequential images, outputting a new bounding box label for each image that runs through our method. The determination of the location of the bounding box on new images is predicted by the network, not a set of interpolation calculations.

## II. Background

CNN's have demonstrated outstanding performance on a wide range of computer vision tasks, such as image classification [8–10], image segmentation [11-12], and object detection [13–15]. We aim to take advantage of these advances for our tracking problem.

Several works have been published on single-target object tracking using convolutional neural networks, like [16] which uses both online and offline training for tracking. In the area of offline training, the authors trained a stacked denoising autoencoder using auxiliary natural images to learn generic image features which are robust against variations. During online training, the output of the encoder is fed into a classification neural network as a feature extractor and an additional classification layer. Others like [17] propose a tracking algorithm by learning a discriminative saliency map using a pre-trained CNN and an online SVM [18]. The SVM discriminates targets from the background by learning target-specific information from the CNN features. Computing the saliency map by back-projecting the CNN features with the guidance of SVM. [19-20] uses a VGGNet model pre-trained with an ImageNet dataset to create a tracking algorithm. Convolutional layers at different levels are jointly used to create the tracking algorithm. The authors of [21] propose a multi-domain CNN for visual tracking. In this work, the network is composed of shared layers and multiple branches of domain-specific layers. The network is pre-trained offline and the fully connected layers including the domain-specific layers are fine-tuned online.

Traditional modern object trackers use a combination of offline and online training approaches for single-target object tracking [16-17], [19–21]. However, our approach is slightly different since we are using the offline training approach presented by [2] with no online training required after the network is trained offline.

### III. DATASET USED

Our SDO AIA image data is collected in two steps: first, event records are downloaded and filtered, second: from the filtered event records, solar images are downloaded or extracted from [22] and annotation files are created.

#### A. Event Record Collection

We extracted all the event reports from the Heliophysics Event Knowledgebase (HEK) during the years 2014 to 2017 for Active Region and Coronal Holes events. We selected the reports from the Helioseismic and Magnetic Imager (HMI) instrument [23] for active regions, and SPoCA module [24] for coronal holes. As a pre-filtering step, we excluded the event reports which had a bounding box outside of the solar limb. In order to do this, we downloaded the image from the closest time when the event started and extracted the value of the sun's radius from the downloaded image's header. Every solar event has its unique specific id which is included in the event report. We grouped event reports using this specific id and saved them on a CSV formatted file. Short events with less than 3 event records are excluded from the dataset.

#### B. Image Matching or Download

For each solar event, we consolidate all of its event reports from HEK in a CSV file which is sorted by the event start time. Every event report has a start time and an end time. For the images we did not find in [22], we downloaded them from Helioviewer [25], calculated the time interval from event start time and end time, and divided the time interval into four equal time intervals. Then we started downloading images from the event start time and kept incrementing the time with the time interval and downloads images until we reached the event end time for each event report.

The data had some anomalies, for example, sometimes certain images from a certain date and time were missing or the downloaded images were entirely black. We took consideration of these problems during the creation of the dataset and removed those events from it to avoid incorrect data points.

#### C. Event labeling on image data

We used two different types of data annotations, the bounding box from HEK and a generated maximal internal box from the chain codes available. Every event report has its bounding box location in the Helioprojective-Cartesian (HPC) coordinate system. To convert the HPC coordinate to the pixel coordinate system, we downloaded the closest available image based on event start time and used the CDELT and CRPIX value from the original JP2 image header for converting to pixel coordinate system. Each event has its annotation file with the bounding box information for every event report and its corresponding image. The annotation starts with the frame number of the video and its bounding box information such as $i$, $x_1, x_2, x_3, x_4, y_1, y_2, y_3, y_4$ whereas $i$ is the frame number of the video on the event folder and the remaining are the bounding box coordinates.

### IV. TRACKING METHODOLOGY

In our experimental section, we are using the approach of Generic Object Tracking Using Regression Networks (GOTURN) [2] to track solar events. This tracking method has been originally designed to be trained offline with the video sequences and to find the generic relationship between appearance and motion. At evaluation/run time it outputs the location of the tracked object within each frame.

#### A. Network Architecture

A custom CaffeNet [26] architecture is used in this tracking method. A diagram is found in Figure 1. The network has three inputs: current image, target image, and bounding box of the initial object location. There are two branches for these two images in the network. The image of the target object and the search image of the search region are fed into a five-layer convolution layer of the CaffeNet architecture. The output of these two branches is then concatenated into a single vector. This vector is then fed into 3 fully connected layers. Each fully connected layer has 4,096 nodes.

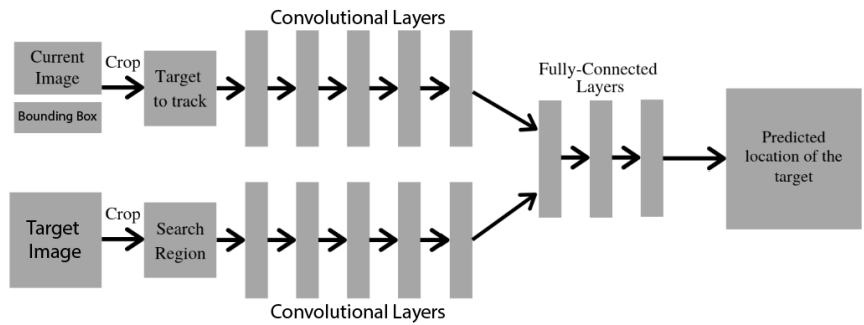

Fig. 1. Network configuration layout

The output of the last fully connected layer is then fed into a four node fully connected layer which represents the bounding box location of the target object. In this network, there is another branch, where the current bounding box is flattened and compared with the output bounding box to calculate the loss of the model. All of the network hyper-parameters are taken from the CaffeNet architecture.

### B. Offline training process

During training, a pair of subsequent frames which have the bounding box location of the object is cropped using the Laplace distribution and then fed into the convolutional layers. The output of these convolutional layers is a high-level representation of the image. This high-level representation of the image is then fed into the fully connected layers which output the location of the target object where the target object has moved from the previous frame.

## V. EXPERIMENTAL EVALUATION

Using data from 2012 to 2018, we have completely separate training and testing datasets. This will clearly allow us to see if our trained network generalizes well in completely unseen data.

### A. Training set

As shown in Table 1, our training set is divided into three parts, using the HEK-reported bounding box coordinates. Note that in these datasets, approximately every 5th image has a truth label for the location of the solar events as produced by the corresponding instrument/model. We used data from 2012 to 2017 only.

TABLE I. TRAIN DATASET DETAILS

| Dataset Name | Total Images | Annotated Images | Total Sequences |
|---|---|---|---|
| AR | 468,402 | 94,680 | 2,872 |
| CH | 128,033 | 25,969 | 1,347 |
| AR & CH | 596,435 | 120,649 | 4,219 |

### B. Test set

We divided our test into three parts. The first test set has solar image data from the year 2018 for AR and CH images and has a total of 19,759 images from 196 videos. The number of annotated frames is 3,994 images for this test set. We split the first test set into two parts based on their event type. The second and third test set have AR and CH events data respectively. The second test set has 2,132 annotated frames from 10,621 images and all of these images are from 76 videos. The third test has a total of 9,138 images from 120 videos. This set has 1,862 annotated frames. As previously mentioned, approximately every 5th image is annotated with the bounding box location of an event.

TABLE II. TEST DATASET AT GLANCE

| Dataset Name | Total Images | Annotated Images | Total Sequences |
|---|---|---|---|
| AR | 10,621 | 2,132 | 76 |
| CH | 9,138 | 1,862 | 120 |
| AR & CH | 19,759 | 3,994 | 196 |

### C. Evaluation Metrics

In order to better capture and evaluate the performance of the proposed tracking method, we are using six different metrics that will capture not only overlap between bounding boxes, but also characteristics like size of bounding boxes and other statistical performance metrics.

*1) Intersection over Union (IoU):* This metric evaluates the intersection over union, measuring the similarity of the predicted bounding box to the ground truth bounding box. The formula for Intersection over Union is shown in equation 1. $GT^i$ denotes the ground truth bounding box for frame $i$ and $T^i$ denotes the predicted bounding box for frame $i$. IoU compares the area of two bounding boxes by the ratio of their shared area and their combined area.

$$\text{IoU}(GT^i, T^i) = \frac{|GT^i \cap T^i|}{|GT^i \cup T^i|} \quad (1)$$

*2) F-score:* The F-score measures the test set accuracy, which can be interpreted as a weighted average of precision and recall. The formulas for precision (2), recall (3) and F-score (4) are as follows.

$$\text{precision} = \frac{n_{tp}}{n_{tp} + n_{fp}} \quad (2)$$

$$\text{recall} = \frac{n_{tp}}{n_{tp} + n_{fn}} \quad (3)$$

$$\text{F-score} = 2 \times \frac{\text{precision} \times \text{recall}}{\text{precision} + \text{recall}} \quad (4)$$

On equations 2 and 3, $n_{tp}$, $n_{fp}$ and $n_{fn}$ denote the total number of true positive tracked boxes, the total number of false-positive tracked box, and the total number of the tracked box by the tracking algorithm respectively.

$$\text{IoU}(GT^i, T^i) \geq 0.5 \quad (5)$$

Note that true positives and false positives are determined by equation (5). This is, when the condition of equation 5 is met, the tracked bounding box is considered to be matched with the ground truth bounding box and it becomes a true positive sample. However, if the condition is not matched, we considered it as a false positive. The condition here is that the Intersection over Union of the boxes is greater than or equal to 0.5. Were 0.5 means that at least half of the box overlaps.

*3) Area Based F1 score (AF1-Score)*: The formula for area-based F1 score [27] is given in Equation (6). In the equation, precision and recall are the ratio of the shared area and the ground truth area, and the ratio of the shared area and the tracked area respectively as shown in Equation (7) and (8).

$$F1 = \frac{1}{\text{no of frames}} \sum_i 2 \frac{\text{precision} * \text{recall}}{\text{precision} + \text{recall}} \quad (6)$$

$$\text{precision} = \frac{|GT^i \cap T^i|}{|GT^i|} \quad (7)$$

$$\text{recall} = \frac{|GT^i \cap T^i|}{|T^i|} \quad (8)$$

Area-based F1 score evaluates the average area coverage of the ground truth bounding box and the tracked bounding box.

*4) Object Tracking Accuracy (OTA):* Object tracking accuracy evaluates the overlaps between the predicted bounding box and the ground truth bounding box. The formula for OTA is shown in Equation (9). For single object tracking, $g^i$ denotes the number of ground truth bounding box in frame *i* which is either a 0 or 1.

$$OTA = 1 - \frac{\sum_i (n_{fn}^i + n_{fp}^i)}{\sum_i g^i} \quad (9)$$

*5) Intersection over Ground Truth (IoGT):* Since intersection over union does not consider if the tracked area fully overlaps with the ground truth area or if the tracked bounding box is bigger than the ground truth box, we will use the intersection over ground truth area. In the scenario when the tracked box covers the entirely the ground truth box and goes considerably beyond its limits, this metric penalizes its score. Ground truth overlaps metrics consider the above circumstance of the tracked box and the formula is given in equation (10).

$$IoGT(GT^i, T^i) = \frac{|GT^i \cap T^i|}{|GT^i|} \quad (10)$$

IoGT metric evaluates the ratio between the shared area and the area of the ground truth box. The maximum value and minimum value for this metric is 1 and 0 respectively. Therefore, if the value is 1, the predicted bounding box covers the entire ground truth box and 0, otherwise.

*6) Average Tracked Box Size (ATB):* Average tracked box size evaluates how much bigger is the size of the predicted box with regards to the ground truth box. This metric is calculated by equation (11).

$$ATB(GT^i, T^i) = \frac{|T^i|}{|GT^i|} \quad (11)$$

$T^i$ and $GT^i$ denotes the area of the tracked bounding box and ground truth bounding box for frame *i*.

## VI. EXPERIMENTAL EVALUATION AND DISCUSSION

We present our experimental evaluation by separating the three different tracking models we trained: 1) Using only Active Regions (**mAR**), 2) Using only Coronal Holes (**mCH**), 3) Mixing both Active Regions and Coronal Holes (**mAR-CH**). Each of our models is trained for 200,000 iterations. We have used Intersection over union (IoU) and Intersection over Ground Truth (IoGT) area metrics to calculate the F-score separately for each of these metrics. We set a threshold of 0.50 for a prediction to be True Positive and value less than 0.50 is considered as False Positive. During the training, in every 2000 iterations we have saved the model weights. We have run the training for 200,000 iterations. Therefore, we have 100 saved model weights for each model. After the training has been completed, we run the test dataset on these saved models to calculate the score for evaluation metrics.

### A. Active Region Results

In our model only trained with the purpose of tracking Active Regions, we achieve the highest performance for all our metrics. This finding is not surprising, however, we are pleased that the difference between the model trained for both events is not extreme, but considerable enough to degrade performance. The mean score of IoGT area is 0.5495, which indicated that at least half of the bounding boxes (predicted vs. ground truth) overlap, something that is also reflected on the Fscore (IoGT). One of the most surprising results found is that for the AR model only the average size of the tracked box (ATB) is 1.6461 times the ground truth box. This means we have predicted larger boxes that might be capturing a wider amount of pixel differences between images. Table 3 shows all the values for the calculated metrics.

TABLE III. PREDICTION OF ACTIVE REGION BOUNDING BOXES RESULTS

| Model | IoU | F-score (IoU) | AF1-Score | OTA | IoGT | F-score (IoGT) | ATB |
|---|---|---|---|---|---|---|---|
| mAR | 0.4297 | 0.4164 | 0.5060 | 0.3987 | 0.5495 | 0.5411 | 1.6461 |
| mAR-CH | 0.3728 | 0.3118 | 0.4521 | 0.2976 | 0.4581 | 0.4064 | 0.8783 |

With respect to the training iterations, Figure 2 shows a plot of our evaluation metric values calculated per iteration. The purpose of these plots is to determine the stability of our model after a certain number of iterations. As we can see after around 100,000 iterations, the variability of our metrics is reduced.

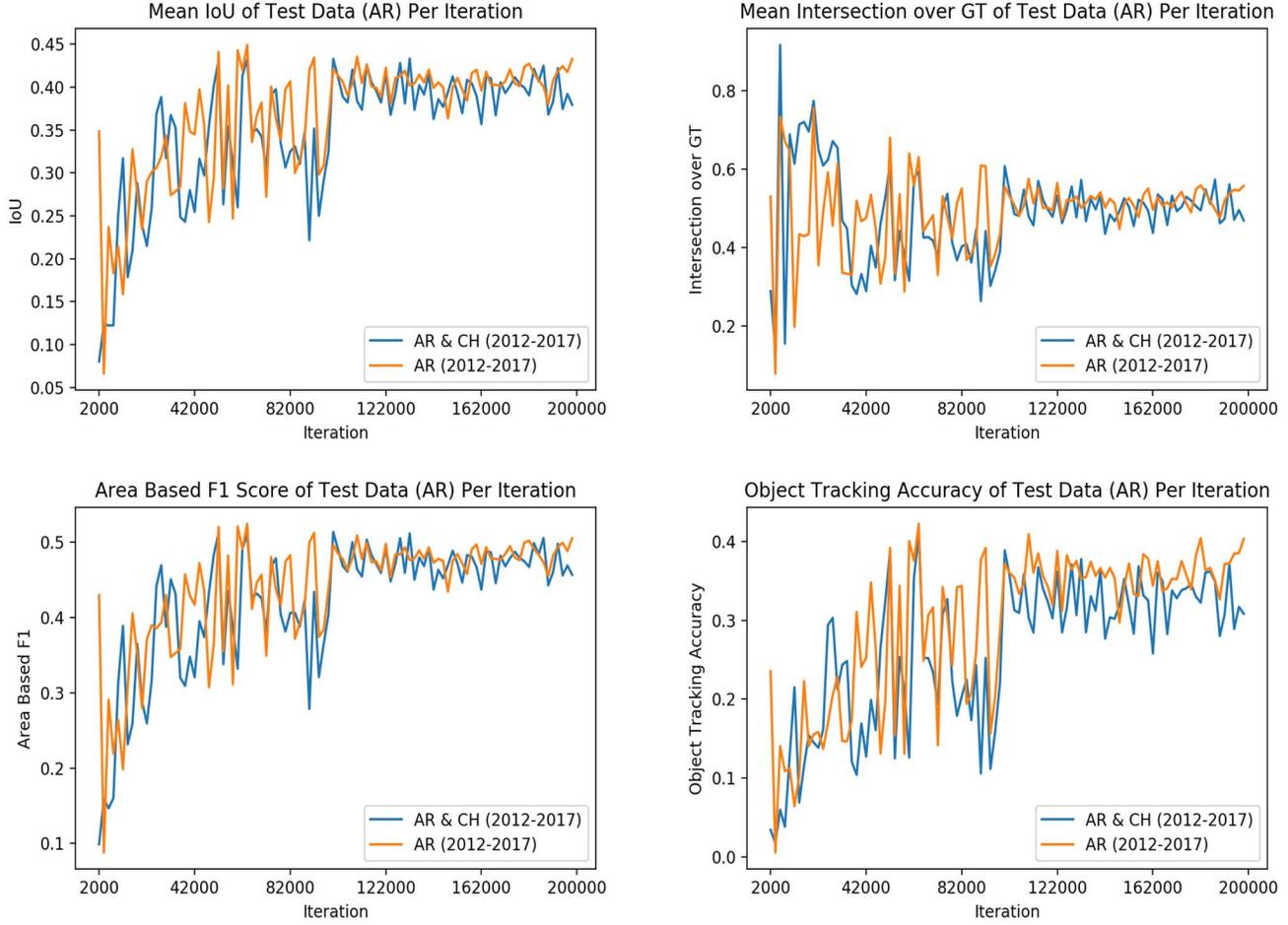

Fig 2. Plots of IoU, IoGT, Area Based F1 score and OTA scores on the test data for Active Regions.

## B. Coronal Holes Results

Similar to the Active Regions, Table 4 shows that the model only trained with Coronal Holes performs better than the one with both events. In terms of how much larger the predicted boxes and the labels are, we see that they are not considerably bigger, meaning our original labels do a better job at containing the Coronal Holes than the Active regions.

TABLE IV. PREDICTION OF CORONAL HOLE BOUNDING BOXES RESULTS

| Model | IoU Mean | F-score (IoU) | AF1-Score | OTA | IoGT Mean | F-score (IoGT) | ATB |
|---|---|---|---|---|---|---|---|
| **mCH** | **0.4156** | **0.3815** | **0.4535** | **0.3494** | **0.5739** | **0.5586** | **1.1172** |
| mAR-CH | 0.3396 | 0.2658 | 0.3721 | 0.2395 | 0.4776 | 0.4309 | 0.9411 |

Both Table 4 and Figure 3 show that the difference in performance between models is more marked than for Active Regions. We again see the model stabilizing after 100K iterations, but we do see some interesting results for the IoGT and OTA metrics as it shows improved performance in the first iterations and wild changes between them. This could be in part because the predicted labels are closer to the ground truth, but not consistently through the different iterations.

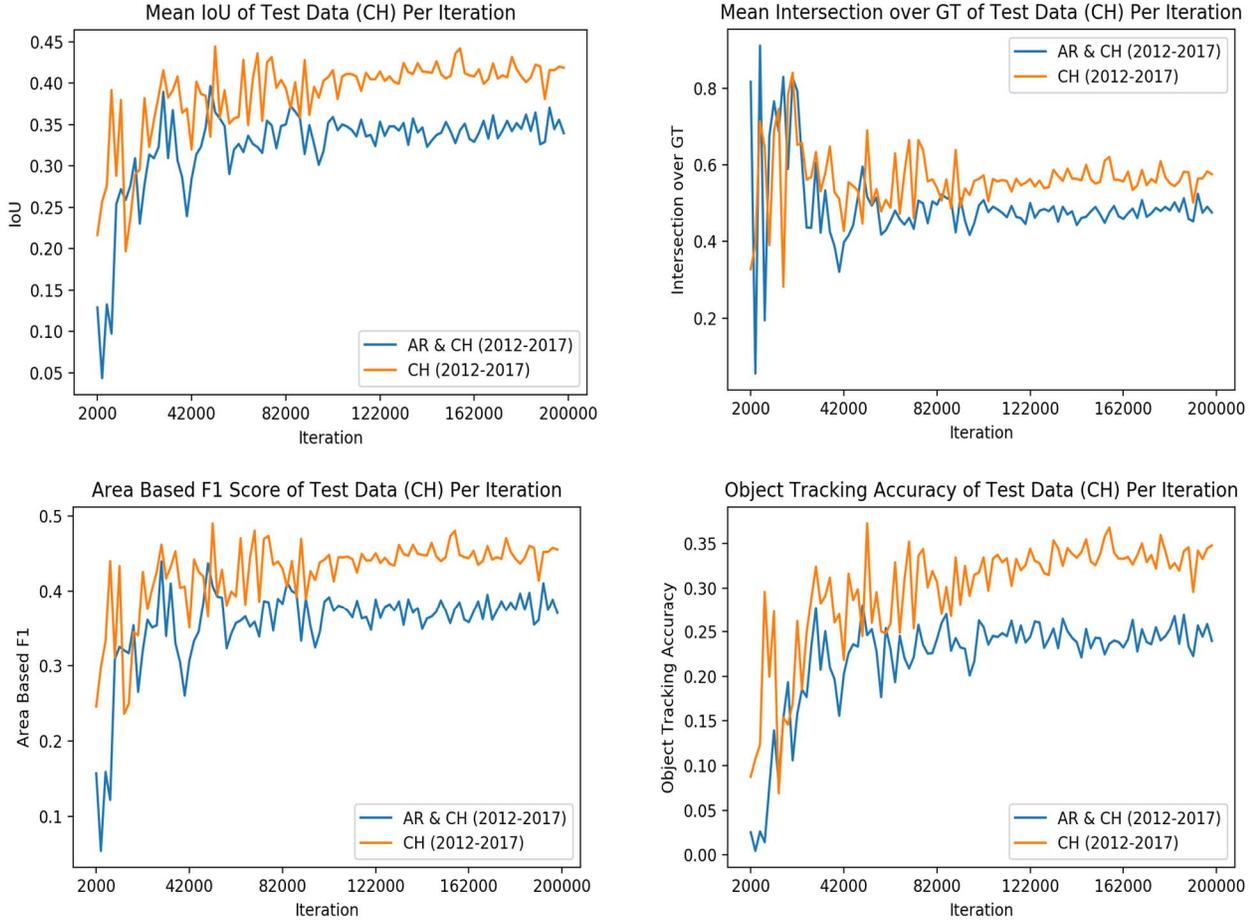

Fig. 3. Plots of IoU, IoGT, Area Based F1 score and OTA scores on the test data for Coronal Holes.

## C. Active Region and Coronal Hole Results

In order to evaluate a model that could potentially perform multiple event tracking at once, we trained and tested on all AR and CH event data put together. It seems like the two events and their image signatures confuse the model enough to bar it from distinguishing easily. This is clearly indicated by producing smaller bounding boxes (ATB), a result that is more representative on Coronal Holes (Table 4), but not for Active Regions (Table 3). While this result might seem discouraging, we believe that once we have more events to identify in a single model, these should improve considerably. This could be an artifact of the current event pairing we had for this evaluation. Table 5 presents the results for these models.

TABLE V.   AR & CH RESULTS

| Model | IoU Mean | F-score (IoU) | AF1-Score | OTA | IoGT Mean | F-score (IoGT) | ATB |
|---|---|---|---|---|---|---|---|
| mAR-CH | 0.3525 | 0.2836 | 0.4031 | 0.262 | 0.47 | 0.4214 | 0.9167 |

## D. How to interpret these results

While not achieving over 50% on most overlap metrics is intuitively weak, in Active Regions our method takes a very big performance hit when events are on the limb of the sun, briging down the metrics averages. It is very tricky to evaluate tracking results when there are no available full tracking labels. While we are aware that there are interpolated tracks available, we wanted to only compare with FFT modules and instruments. It is not possible to show a video of a tracked sequence, Figure 5 shows an example of how our method performs compared to available labels and demonstrates that limb issue with lows scores. The Figure showcases how the predicted boxes (red) overlap with the HEK event labels (white). It is clear to see that the tracking is clearly following the solar event, but it becomes subjective when compared against the event label, as both are visually correct. We theorize that our method's performance is still very good, but currently we do not have a proper way of evaluating it completely. The final section of this paper features some ideas about this.

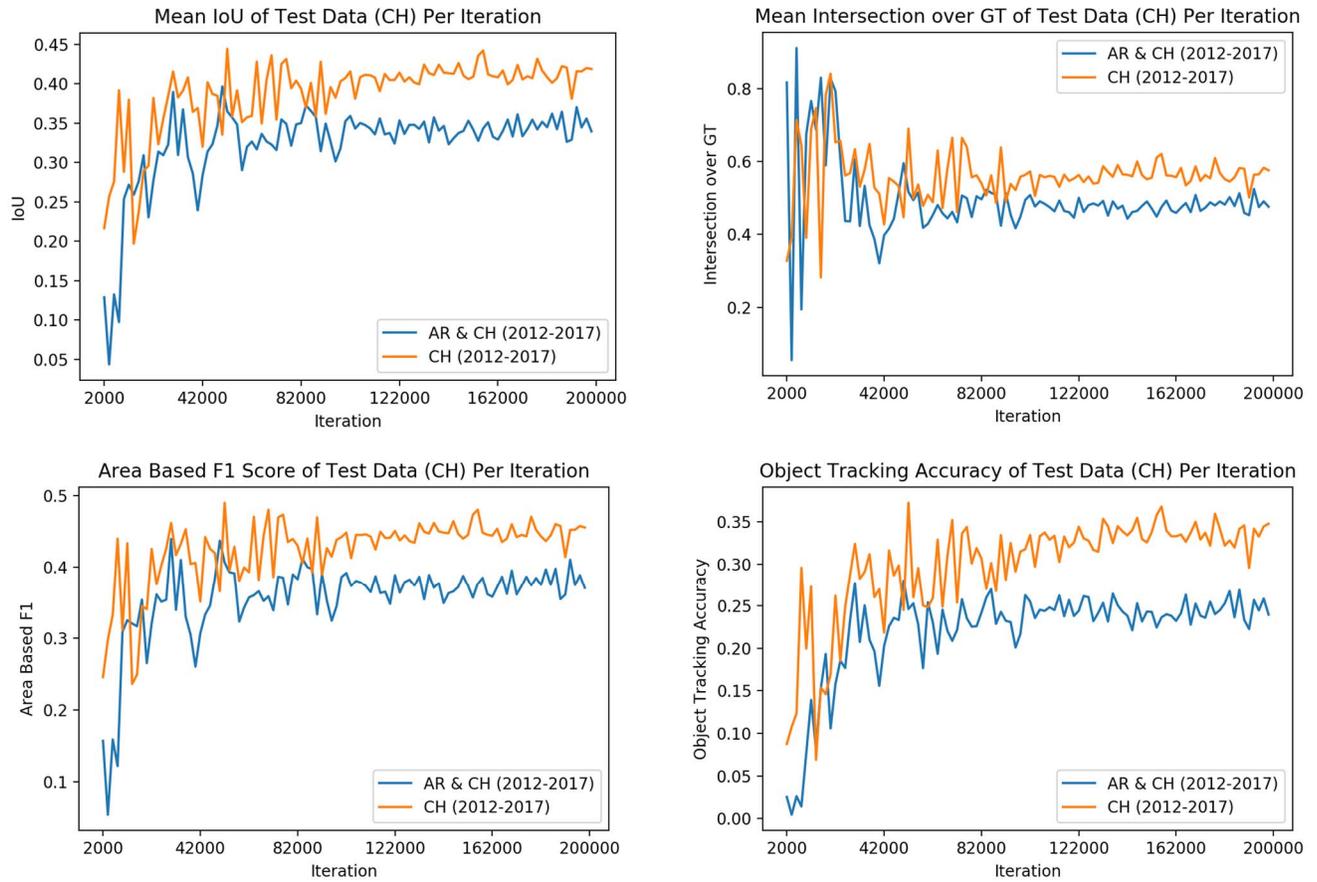

Fig. 4. Line graphs of IoU, IoGT, Area Based F1 score and OTA score of Test Data (Active Region & Coronal Holes)

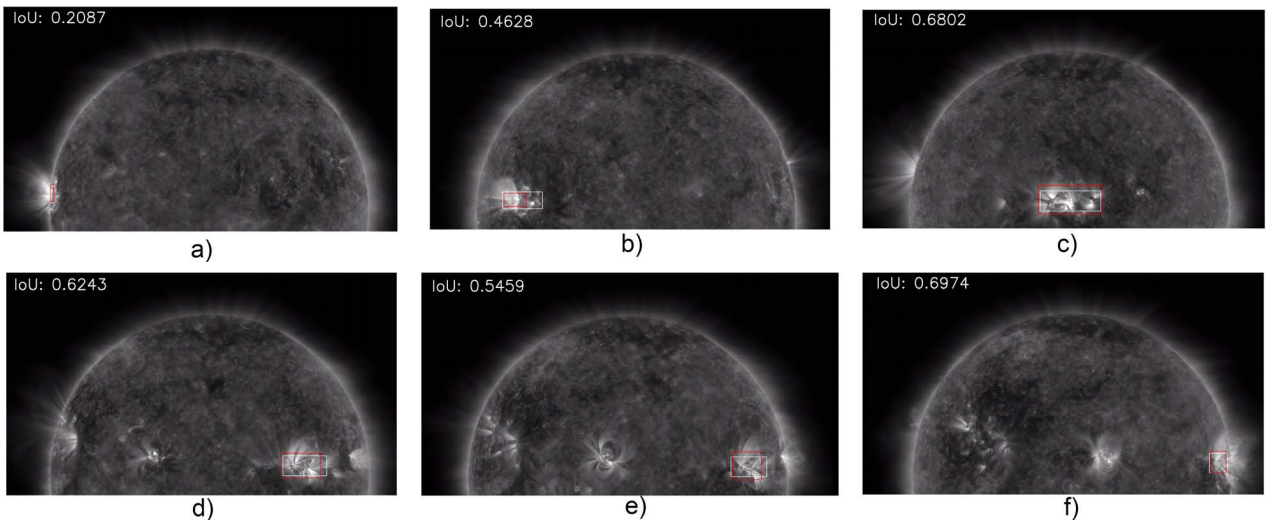

Fig. 5. Tracking steps for a single AR event. Each image compares the predicted label (red) with the HEK label (white).

## VII. CONCLUSION AND FUTURE WORK

In this work we have demonstrated the feasibility of creating a solar event tracking method using Generic Object Tracking Using Regression Networks [2]. With the purpose of creating continuously labeled event trajectories for the enhancement of labeled image data within the SDO mission, we have trained all models using 200,000 iterations to make sure the model is refined and stable as shown in Figures 2 and 3. The scope of our work, for now, is limited to only two different types of solar events, in scenarios where we have one model for each and a single model to track two different events at the same time. While the results of our evaluation metrics are not very promising, we believe they show potential in the sense that we are not missing the intermediate bounding boxes (available module/instrument generated labels) by much. It has to be considered as well that this approach is purely visual and uses the images and no other metadata from them to generate event tracks. This approach is purely based on image data and a starting label, which has the promise of functioning on any other type of solar imagery and can use non-professional labels as a starting point. The generated labels provide the continuous data that label-hungry deep learning algorithms can use to learn and represent the fine grained movements of solar phenomena for a plethora of other applications. We opted to use a combined model since our original scope for this project was to deploy the model on an Intel Movidius neural compute stick for portability. Having one model would simplify the deployment.

As for future work, we are currently working on adding several other types of solar events to our tracking method, as well as creating an ensemble of models to compare against the single model approach. We will expand our evaluation to use interpolated labels [7] instead of only using module/instrument generated labels to observe and verify performance and compare against non-deep learning methods.

## REFERENCES

[1] A. Dosovitskiy et al., "Flownet: Learning optical flow with convolutional networks," in Proceedings of the IEEE international conference on computer vision, 2015, pp. 2758–2766.

[2] D. Held, S. Thrun, and S. Savarese, "Learning to Track at 100 FPS with Deep Regression Networks," in Computer Vision – ECCV 2016, 2016, pp. 749–765.

[3] B. Babenko, M.-H. Yang, and S. Belongie, "Robust Object Tracking with Online Multiple Instance Learning," IEEE Trans. Pattern Anal. Mach. Intell., vol. 33, no. 8, pp. 1619–1632, Aug. 2011.

[4] J. Gao, H. Ling, W. Hu, and J. Xing, "Transfer Learning Based Visual Tracking with Gaussian Processes Regression," in Computer Vision – ECCV 2014, 2014, pp. 188–203.

[5] P. C. H. Martens et al., "Computer Vision for the Solar Dynamics Observatory (SDO)," Solar Phys., vol. 275, no. 1, pp. 79–113, Jan. 2012.

[6] S. F. Boubrahimi, B. Aydin, M. Schuh, D. Kempton, and R. Ma, "Spatiotemporal Interpolation Methods for Solar Event Trajectories," vol. 236, no. 1), May 2018.

[7] S. F. Boubrahimi, B. Aydin, D. Kempton, S. S. Mahajan, and R. Angryk, "Filling the Gaps in Solar Big Data: Interpolation of Solar Filament Event Instances," in 2016 IEEE International Conferences on Big Data and Cloud Computing (BDCloud), Social Computing and Networking (SocialCom), Sustainable Computing and Communications (SustainCom) (BDCloud-SocialCom-SustainCom), 2016, pp. 97–104.

[8] K. Simonyan and A. Zisserman, "Very Deep Convolutional Networks for Large-Scale Image Recognition," arXiv [cs.CV], 04-Sep-2014.

[9] C. Szegedy et al., "Going Deeper with Convolutions," arXiv [cs.CV], 17-Sep-2014.

[10] K. He, X. Zhang, S. Ren, and J. Sun, "Deep Residual Learning for Image Recognition," 2016 IEEE Conference on Computer Vision and Pattern Recognition (CVPR). 2016.

[11] H. Noh, S. Hong, and B. Han, "Learning Deconvolution Network for Semantic Segmentation," 2015 IEEE International Conference on Computer Vision (ICCV). 2015.

[12] J. Long, E. Shelhamer, and T. Darrell, "Fully convolutional networks for semantic segmentation," 2015 IEEE Conference on Computer Vision and Pattern Recognition (CVPR). 2015.

[13] R. Girshick, J. Donahue, T. Darrell, and J. Malik, "Rich Feature Hierarchies for Accurate Object Detection and Semantic Segmentation," 2014 IEEE Conference on Computer Vision and Pattern Recognition. 2014.

[14] R. Girshick, "Fast R-CNN," 2015 IEEE International Conference on Computer Vision (ICCV). 2015.

[15] S. Ren, K. He, R. Girshick, and J. Sun, "Faster R-CNN: Towards Real-Time Object Detection with Region Proposal Networks," IEEE Trans. Pattern Anal. Mach. Intell., vol. 39, no. 6, pp. 1137–1149, Jun. 2017.

[16] N. Wang and D.-Y. Yeung, "Learning a Deep Compact Image Representation for Visual Tracking," in Advances in Neural Information Processing Systems 26, C. J. C. Burges, L. Bottou, M. Welling, Z. Ghahramani, and K. Q. Weinberger, Eds. Curran Associates, Inc., 2013, pp. 809–817.

[17] S. Hong, T. You, S. Kwak, and B. Han, "Online Tracking by Learning Discriminative Saliency Map with Convolutional Neural Network," in International Conference on Machine Learning, 2015, pp. 597–606.

[18] C. Cortes and V. Vapnik, "Support-vector networks," Mach. Learn., vol. 20, no. 3, pp. 273–297, Sep. 1995.

[19] L. Wang, W. Ouyang, X. Wang, and H. Lu, "Visual tracking with fully convolutional

[20] C. Ma, J.-B. Huang, X. Yang, and M.-H. Yang, "Hierarchical convolutional features for visual tracking," in Proceedings of the IEEE international conference on computer vision, 2015, pp. 3074–3082.

[21] H. Nam and B. Han, "Learning multi-domain convolutional neural networks for visual tracking," in Proceedings of the IEEE Conference on Computer Vision and Pattern Recognition, 2016, pp. 4293–4302.

[22] A. Kucuk, J. M. Banda, and R. A. Angryk, "A large-scale solar dynamics observatory image dataset for computer vision applications," Sci Data, vol. 4, p. 170096, Jul. 2017.

[23] J. Schou et al., "Design and Ground Calibration of the Helioseismic and Magnetic Imager (HMI) Instrument on the Solar Dynamics Observatory (SDO)," Solar Phys., vol. 275, no. 1, pp. 229–259, Jan. 2012.

[24] C. Verbeeck, V. Delouille, B. Mampaey, and R. De Visscher, "The SPoCA-suite: Software for extraction, characterization, and tracking of active regions and coronal holes in EUV images," Astron. Astrophys. Suppl. Ser., vol. 561, p. A29, Jan. 2014.

[25] J. Ireland, K. Hughitt, D. Müller, G. Dimitoglou, P. Schmiedel, and B. Fleck, "The Helioviewer Project: Discovery For Everyone Everywhere," presented at the AAS/Solar Physics Division Meeting #40, 2009.

[26] Y. Jia et al., "Caffe: Convolutional Architecture for Fast Feature Embedding," in Proceedings of the 22nd ACM international conference on Multimedia, 2014, pp. 675–678.

[27] A. W. M. Smeulders, D. M. Chu, R. Cucchiara, S. Calderara, A. Dehghan, and M. Shah, "Visual Tracking: An Experimental Survey," IEEE Trans. Pattern Anal. Mach. Intell., vol. 36, no. 7, pp. 1442–1468, Jul. 2014.